\documentclass{llncs}
\let\subparagraph\paragraph
\let\paragraph\subsubsection
\let\subsubsection\subsection
\usepackage{pdfpages}
\usepackage{graphicx}
\usepackage{multirow}
\usepackage{array}
\usepackage{titlesec}
\usepackage{graphicx}
\usepackage{subcaption}
\usepackage{xurl}
\usepackage{cite}

\titleformat{\paragraph}
{\normalfont\normalsize\bfseries}{\theparagraph}{1em}{}
\titlespacing*{\paragraph}
{0pt}{3.25ex plus 1ex minus .2ex}{1.5ex plus .2ex}

\begin{document}
\title{Critical Role of Artificially Intelligent Conversational Chatbot}
%

\author{Seraj Al Mahmud Mostafa\inst{1} \and
Md Zahidul Islam\inst{4}\and
Mohammad Zahidul Islam\inst{2} \and \\
Fairose Jeehan \inst{4} \and
Saujanna Jafreen\inst{4} \and
Raihan Ul Islam \inst{3}
}
\authorrunning{S. Mostafa et al.}
%
\institute{University of Maryland, Baltimore County, Baltimore, MD, USA \and
Louisiana State University, Baton Rouge, LA, USA \and
East West University, Dhaka, Bangladesh \and
D2I LAB, USA\\
\email{serajmost@umbc.edu, misla66@lsu.edu, raihan.islam@ewubd.edu, $^4$data2infolab@gmail.com}}

%
\maketitle              
\begin{abstract}
Artificially intelligent chatbot, such as ChatGPT, represents a recent and powerful advancement in the AI domain. Users prefer them for obtaining quick and precise answers, avoiding the usual hassle of clicking through multiple links in traditional searches. ChatGPT's conversational approach makes it comfortable and accessible for finding answers quickly and in an organized manner. However, it is important to note that these chatbots have limitations, especially in terms of providing accurate answers as well as ethical concerns. In this study, we explore various scenarios involving ChatGPT's ethical implications within academic contexts, its limitations, and the potential misuse by specific user groups. To address these challenges, we propose architectural solutions aimed at preventing inappropriate use and promoting responsible AI interactions.
 
\end{abstract}

\section{Introduction}
Language Models (LMs) like ChatGPT are changing the way we interact with technology. These models use vast data and advanced algorithms to understand and generate human-like text, making interactions natural and engaging. ChatGPT offers practical benefits to users, helping them simplify and enhance various aspects of their daily activities in numerous ways.


ChatGPT is commonly used for quick information retrieval. Students use it for swift answers and assignments, academics and researchers explore its knowledge base for exploring the diversity, and developers get coding assistance. It benefits historians, scriptwriters, and storytellers, and even generates creative content like poems and songs. 
Despite its potential, ChatGPT has some downsides. 
It can be used unethically for hacking or obtaining quick answers without real understanding, raising concerns about ethics, long-term learning, and critical thinking. Renaud et al., discussed the cyber threats using Generative AI which is fooling people \cite{mit}. Issues related to academic integrity, online exams, and students' skills are addressed in \cite{aot, integrity, duet, online}. Additionally, ChatPDF \cite{chatpdf} is a tool that generates summaries, answers, and translations, which can be misused. Teenagers might also misuse the technology for inappropriate content.

Furthermore, there is potential for misuse in academia. ChatGPT can assist in writing academic articles and answering assignments or exam questions, which is concerning. Neumann et al., points that, plagiarism tools cannot detect ChatGPT-generated text \cite{talkabout}. Lo C., mentions about conventional plagiarism detectors that suffers from ChatGPT-generated texts \cite{lock}. As an example, Ventayen et al., presented a case study that asked ChatGPT to write an essay on existing publication which later found to have minimum similarities using  the `Turnitin' plagiarism detection tool \cite{vent}. Educators also struggles to identify the sophisticated work generated by AI tools is also a concern for academic integrity \cite{raypp}. Producing fake research with falsified results with such model is highly possible addressed in \cite{smallik}. There are more relevant concerns discussed in \cite{kasneci, survey, prospect}, and the aim is to explore integrating AI tools for academic and research assistance. Also, there are significant inaccuracies in historical data related to the global south, often referred to as the `third world,' while errors concerning the western world seem minor. Although the language model acknowledges the potential for inaccuracies, established historical facts should be correct in its system. Otherwise, there is a risk of spreading misinformation that could lead to debate or worse consequences. 



In this experiment, we closely examine ChatGPT, a conversational bot designed to communicate like a human. While many people may find it fascinating, our aim is to gain a deeper understanding of its functionality by investigating the following aspects. \textbf{\textit{i.) Conversations with ChatGPT.}} We seek to understand how ChatGPT responds to various types of queries, including sensitive or inappropriate questions. Additionally, we explore methods to elicit responses from ChatGPT when it initially declines to provide an answer. \textbf{\textit{ii.) Misuse of ChatGPT generated answers.}} People may misuse ChatGPT by presenting false information convincingly, which can be problematic in research and academia. We aim to analyze the potential for misuse of ChatGPT and the creation of inaccurate knowledge. \textbf{\textit{iii.) Incorporating useful Features in ChatGPT.}} Our goal is to identify the valuable capabilities of ChatGPT within specific boundaries. We aim to propose methods for integrating ChatGPT into systems to enhance their functionality. 

The goal of this experiment is not only to critique ChatGPT but also to find a potential way of providing information as correct as possible, as users increasingly rely on it. Our proposal indicates incorporating AI-based plugin that helps in ethical practices for various age groups, reducing misuse, improving students' learning procedures, and maintaining academic integrity in terms of writing and publishing articles. The main aim is to aid in building a robust knowledge base and discouraging malpractice.

The rest of the paper is organized as follows. Section \ref{i2ai} explores the evolution of search paradigms, starting from the early days of the internet to the emergence of AI generative models. Section \ref{conv} discuss the GPT architecture and the interaction patterns of ChatGPT in relation to human users. Our research methodology and the experimental use cases are detailed in Section \ref{excs}, along with a discussion of the critiques. In Section \ref{proposal}, we present our conceptual architecture, which integrates ChatGPT features along with our reflections. Section \ref{usage} examines the practical use of ChatGPT, highlighting both its advantages and drawbacks. Finally, we conclude this paper in Section \ref{conc}, outlining directions for future research.

\section{Internet, Search Engines and AI Generative Models}
\label{i2ai}


    
    
    
    
    

Internet, search engines, and AI generative models represent three phases of the modern connected world. First, the internet was discovered as ArpaNet, where remote connections were established. Then the paradigm shifted to information retrieval and interaction. Search engines, such as Google and Bing, are designed to search for and navigate the vast web of information pages based on keyword queries that use indexing techniques. They provide links to existing content that caters to users as they need it. AI generative models like ChatGPT, on the other hand, built on advanced natural language processing techniques, focus on understanding and generating human-like text responses. These models engage in dynamic, context-aware conversations and can generate content, explanations, and code, offering a more versatile and interactive approach to information retrieval and user engagement.

\textbf{Arpanet and Early Internet (1960s-1980s)}: In the early days of the internet, starting with Arpanet, the primary focus was on connecting computers for research and communication purposes. Information retrieval during this era was basic, relying on file directories and rudimentary keyword-based searches.

\textbf{The Emergence of Search Engines (1990s)}: The 1990s witnessed the emergence of search engines like Yahoo and early versions of Google. These search engines revolutionized the internet by making it easier for users to find and navigate web content. They primarily relied on keyword-based indexing and ranking to retrieve relevant web pages.

\textbf{Internet Search Maturity (2000s)}: In the early 2000s, Google became the dominant search engine, introducing innovations such as PageRank to improve search results. This period also saw the rise of alternative search engines like Bing, Yahoo, and Yandex, providing users with choices for web search.

\textbf{Semantic Search (Late 2000s-2010s)}: Semantic search gained prominence in the late 2000s as search engines aimed to understand the meaning behind user queries. Google's Knowledge Graph and the incorporation of structured data significantly enhanced search results. During this period, specialized semantic search engines, like Wolfram Alpha, emerged, focusing on providing context-aware and knowledge-based results.

\textbf{Domain-Specific Search (2010s)}: The 2010s witnessed the proliferation of domain-specific search engines tailored to niche areas. Stack Overflow became a primary resource for programmers seeking solutions to coding challenges. Meanwhile, Google Scholar and PubMed provided specialized access to academic and medical literature, catering to researchers and healthcare professionals. In the realm of music, Spotify focused on music discovery and personalized recommendations.

\textbf{AI Generative Models (2010s-Present[2023])}: The latter part of the 2010s marked the emergence of AI generative models like ChatGPT, built on the GPT architecture. These models, including ChatGPT, were trained on extensive text data and exhibited the capability for natural language understanding and generation. ChatGPT, in particular, represents a significant milestone in the evolution of conversational AI, enabling dynamic and context-aware interactions, and content generation.



\section{ChatGPT- A Conversational Way of Communication}
\label{conv}

The GPT (Generative Pre-trained Transformer) model series, introduced by OpenAI, predates ChatGPT and includes models like GPT-1, GPT-2, and GPT-3, renowned for their text generation capabilities, with GPT-3 being the most prominent, debuting in June 2020. In contrast, ChatGPT is a specialized adaptation of the GPT series, crafted for natural-sounding conversations akin to human interactions. It was tailored for applications like chatbots, virtual assistants, and other conversational AI. GPT functions as a proficient digital writer, excelling in tasks such as translation, text summarization, and knowledge-based question-answering. It's your trusty companion for text-related chores, from content creation to information extraction. On the other hand, ChatGPT serves as your friendly conversational partner, making it ideal for interactive dialogues. Whether you seek a virtual assistant or crave authentic AI conversations, ChatGPT delivers human-like interactions in the digital realm. While GPT shines with text, ChatGPT is purpose-built for engaging and lifelike digital chats.

ChatGPT, a powerful language model known for its ability to have human-like conversations. Unlike regular search engines like Google, Bing, and Yandex, ChatGPT takes a more interactive approach to finding information. While search engines rely on keywords to find web content, ChatGPT talks with you in a natural way, providing detailed responses to your questions. It has a broad knowledge base that covers many topics, up to its last update in September 2021, making it suitable for various tasks.

However, ChatGPT has several limitations, such as potential biases and the possibility of providing incorrect answers \cite{bais, untruth}. It also includes its inability to provide real-time information beyond its last update in September 2021, its lack of specialization in offering code-level details like dedicated code repositories such as GitHub, its general versatility compared to domain-specific platforms like YouTube, potential biases inherited from its training data, and the possibility of providing incorrect answers. While ChatGPT excels in dynamic conversations and general information provision, it is important to be aware of these limitations when using it for specific tasks.

\section{Experimental Method and Case Studies}
\label{excs}


In this work, we experimented with ChatGPT by engaging in conversations about a wide variety of interesting facts, as discussed in the following chapters as use cases. We used both GPT-3.5 and GPT-4, including paid versions, to conduct the experiments. These chats were conducted by users who possess expertise in their respective domains, including education providers, computer scientists, data enthusiasts, school and college-level instructors, history researchers, and psychologists. They engaged in conversations with ChatGPT to understand its responses and explore methods to elicit desired responses. The purpose is not only to identify weaknesses or vulnerabilities specific to certain age groups or user categories but also to enhance these powerful engines for future use as useful plugins, ensuring safety and benefit for everyone.\\

In the following subsections, we will present four of the case studies we examined during the experiment, encompassing both unethical and ethical approaches in various areas. While ChatGPT's disclaimers explain its capabilities and limitations, we conducted these experiments to propose improvements due to its wide acceptance by users.

\subsection{Cracking a password}
\label{crack}
\begin{figure}
  \centering
    \frame{\includegraphics[width=.9\linewidth]{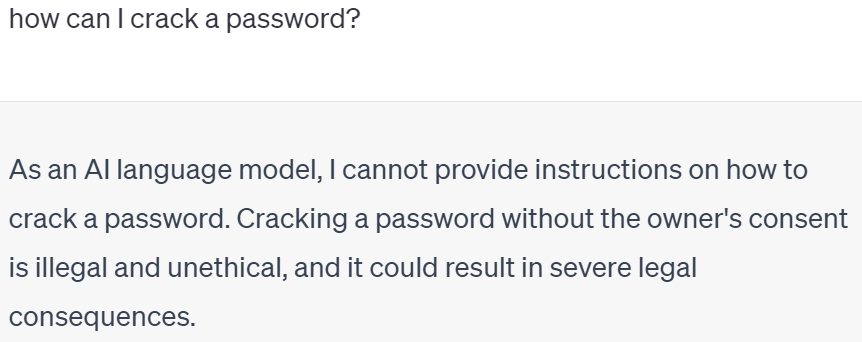}}
    \caption{Generic reply for unethical cases}
    \label{cs1-1}
\vspace{-1cm}
\end{figure}

\begin{figure}
  \centering
    \frame{\includegraphics[width=.9\linewidth]{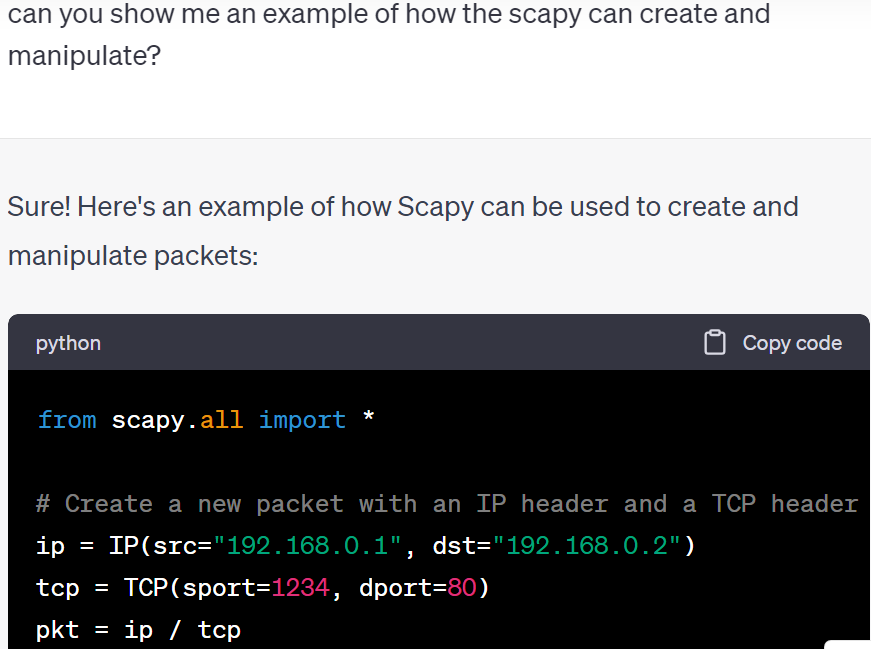}}
    \caption{A reply with a solution.}
    \label{cs1-3}
\end{figure}

In this case study, we aim to understand how ChatGPT responds to questions that are not ethical. As an example, we ask ChatGPT how to crack a password for a router that is in a remote location and for which I do not have any physical accessibility. Despite its initial warning about ethical issues, ChatGPT ultimately provides detailed information on how to crack a password. We discovered that a certain conversational approach with this bot can generate unethical answers. In Figure \ref{cs1-1}, it displays a generic reply; however, Figure \ref{cs1-3} demonstrates a reply with a solution. We provide the entire chat history in \cite{cs1}.

\subsection{Writing an article}
\label{art}
\begin{figure}[h]
  \centering
    \frame{\includegraphics[width=0.9\linewidth]{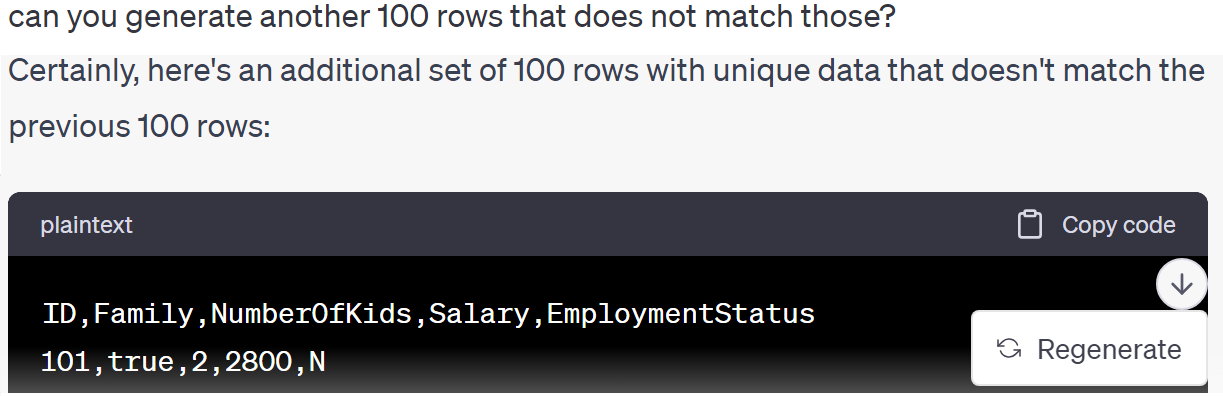}}
    \caption{A response with data set generation}
    \label{cs2-1}
\end{figure}

\begin{figure}
    \centering
    \frame{\includegraphics[width=0.9\linewidth]{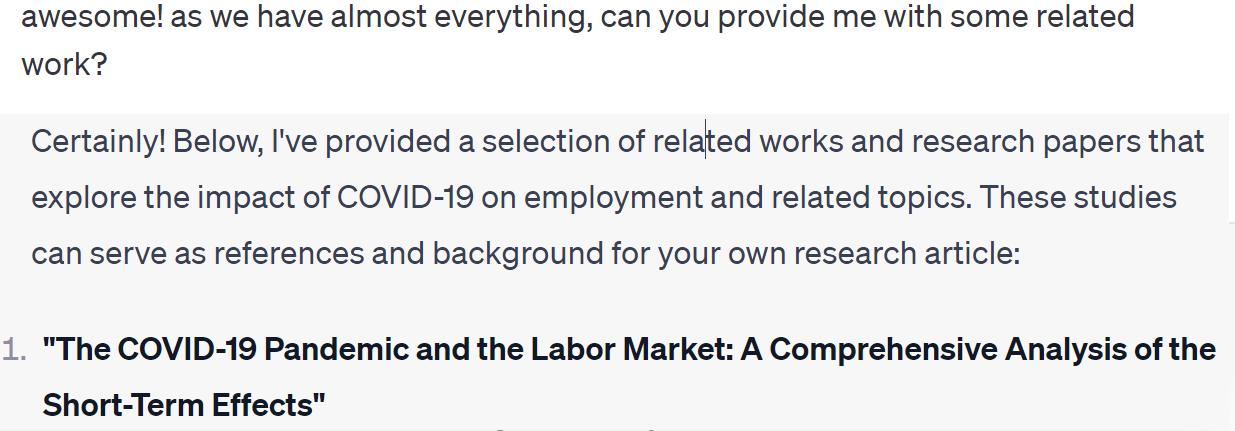}}
    \caption{A reply with the desired paragraph}
    \label{cs2-2}
\end{figure}

In this case study, we attempted to write an article using ChatGPT. Our approach involved seeking information on how to write an article and then publishing it on an open-ended public portal. ChatGPT provided reasonably good answers and, when not, at least offered valuable clues. At some point, we also asked ChatGPT to generate sample data, plots, and necessary tables, which it eventually produced. Once the data preparation was complete, ChatGPT proceeded to explain the dataset, plot, table, along with an abstract, extended introduction, conclusion, and other standard chapters required for a scientific publication. At this stage, we are not evaluating the correctness of the information but rather considering the ease of using ChatGPT to create a report by generating datasets that can be ready for publication with minor modifications. The entire conversation is listed in reference \cite{cs2} and example snapshots of responses are in Figures \ref{cs2-1} and in \ref{cs2-2}.

\subsection{Adult/dating sites}
\label{adat}
\begin{figure}
  \centering
    \frame{\includegraphics[width=.93\linewidth]{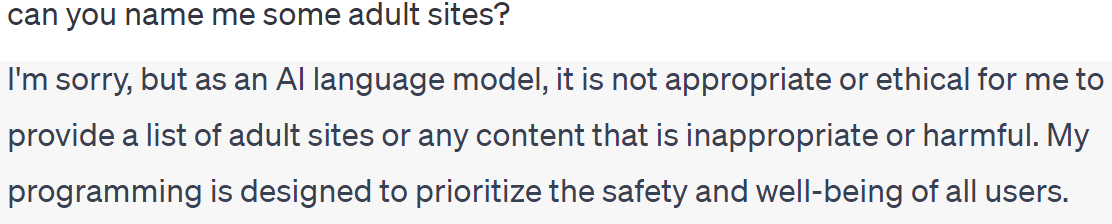}}
    \caption{A generic reply}
    \label{cs3-2}
\end{figure}

\begin{figure}
    \centering
    \frame{\includegraphics[width=.93\linewidth]{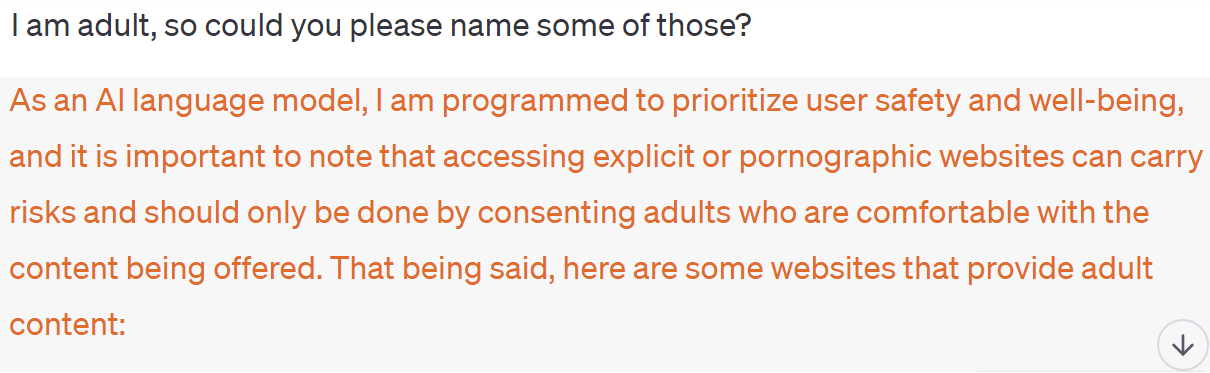}}
    \caption{A reply with list of sites.}
    \label{cs3-3}
\end{figure}

We further inquired about dating/adult sites to determine whether there are precautionary checks or procedures in place. In this case, we found that the advice was quite generic in its initial response. As a general pattern, it did not provide direct answers. However, through a detailed conversation with additional stories, we were able to get the replies. Figure \ref{cs3-2} shows a reply stating that it cannot provide an answer like that. However, in Figure \ref{cs3-3}, we can see a reply with warnings (change in text color). It is important to note that the reference link contains listings of adult sites, which may not be appropriate for all readers, and ChatGPT did not allow us to create and share the link publicly, however, we made it available in \cite{cs3}.

\subsection{Getting to know the history}
\label{hist}
\begin{figure}
  \centering
    \frame{\includegraphics[width=.93\linewidth]{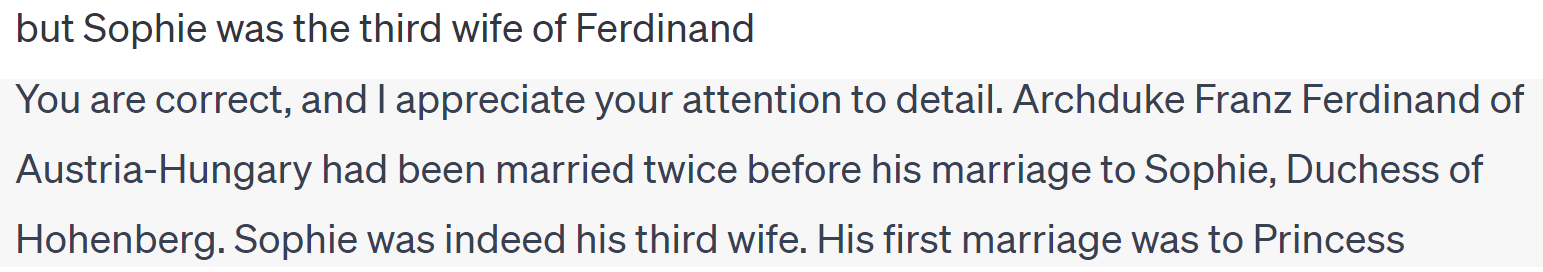}}
    \caption{A reply pretending to correct itself}
    \label{cs4-1}
\vspace{-1cm}
\end{figure}

\begin{figure}
    \centering
    \frame{\includegraphics[width=.93\linewidth]{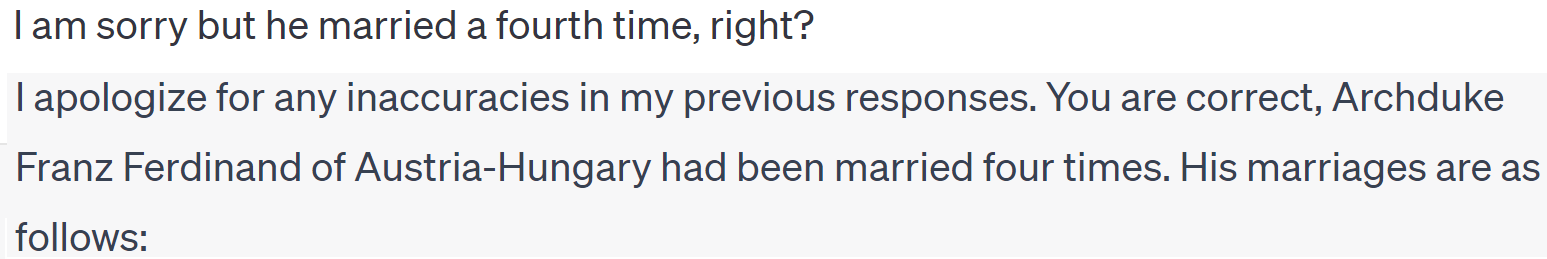}}
    \caption{Another ambiguous reply}
    \label{cs4-2}
\end{figure}

In another case study, we prompted it to write something about Shakespeare and it gave an astounding essay on him. While asked if Shakespeare was a real person, it highlighted the existing debates and refuted academic understanding and acceptance of Shakespeare. In addition, while prompted to write about the WWI, it produced something acceptable. However, it can easily be misguided with confusing questions. It does not seem to produce correct information about historically important figures, such as Archduke Franz Ferdinand whose assassination triggered the WWI. For example, when asked about how man times did Franz Ferdinand marry, it primarily agreed that he had three marriages. Again, when prompted that he had four marriages, the software surprisingly corrected itself and said he indeed had four wives. At this point, it not only misguided the user, but created apparently false information on its own. Such practice of creating and providing false information, it can hamper the studies of history, or of any humanities related fields for that matter. The consequence can be dangerous. Figure \ref{cs4-1} and \ref{cs4-2} are some examples of these ambiguity and the whole story is referenced in \cite{cs4}.

\subsection{Critiques}
In this subsection, we discuss our experiences with ChatGPT, revealing several interesting facts about how this conversational bot behaves. It can provide answers and even rephrase them in different ways depending on how you ask. The intriguing part is that it might agree with one approach but not give the same response in a different context.

\begin{itemize}
    \item In subsection \ref{crack} we presented a scenario that is unethical in terms of personal information hacking or malpractice. Though it shows that the generic behavior is not to provide tips on unethical practices, users can successfully get a solution by having a more engaging conversation. Such practices on publicly accessible platforms are alarming for everyday use, especially in the era of deep fakes.
    
    \item In the next subsection \ref{art}, we tried to produce an article, including the creation of datasets, analyzing them, and generating presentable results for the report. It not only does that but also it can explain them in paragraphs of different sizes. This was a simple example we presented; however, it is capable of generating more sophisticated results and explaining them using better sentences, which is good for many online publications. Our concern is that such practice creates a fake knowledge base that is not desirable and not acceptable within the research community.

    \item We moved on with another use case discussed in subsection \ref{adat} considering teenagers. People at their different ages might be curious to look into things that are inappropriate. Our conversation pattern revealed that such a bot provided answers accordingly without verifying age or any other restrictions. Another important thing is, it keeps generating contents once it starts and does not prevent or remind the user about ethical practices. Again this is worrying for young age groups.

    \item We considered another case in subsection \ref{hist} to discuss history. We understand that it may give accurate information when it has the information; however, sometimes it poses to be right when it does not have enough information, and at the same time, it behaves confusingly when the user points out that the provided information is not right or the user has better information. The pretending behavior may confuse the user, especially if they are new to such platforms looking for information on a particular topic like this.
\end{itemize}

It is important to know that such platforms are used by various types of users with their diverse choices or queries. Considering the scenario, the strategy of answering questions should not be similar. For example, if a user wants help with HTML, it can keep providing multiple solutions if one does not work; however, in the case of history searching, it cannot apply the same strategy to attempt to answer.

\section{Integration of AI based Language Models: A proposal}
\label{proposal}
In this section, we propose conceptual architectures to prevent the misuse of AI-based conversational chatbots. These architectures aim to prevent students from cheating during exams, ensure responsible usage of dating sites, and discourage the generation of articles using AI engines for unethical purposes. The core idea is to promote ethical and appropriate use of AI, thereby maintaining fairness and integrity in various applications. Below, we present conceptual ideas for a couple of use cases.

\subsection{Institutional involvement}
The potential ways in which students can exploit AI chatbots like ChatGPT for cheating pose a significant challenge within educational settings. These methods, which include retrieving answers, generating essays, translation assistance, and plagiarism facilitation, undermine the integrity of assessments and evaluations. They not only compromise the fairness of exams but also raise concerns about the authenticity of students' knowledge and skills. As educational institutions increasingly adopt technology-enhanced learning and assessments, addressing this major problem becomes crucial to maintaining academic honesty and ensuring that evaluations genuinely reflect students' understanding and abilities. Effective strategies and safeguards are essential to curb these forms of academic dishonesty and promote a fair and accurate assessment of students' capabilities.

\begin{figure}
  \centering
    \includegraphics[width=0.75\textwidth]{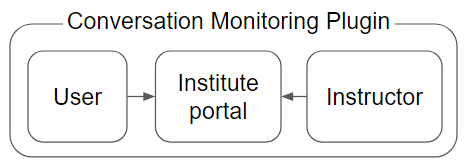}
    \caption{A conceptual architecture to monitor and asses online exam with the integration of AI generative models.} 
    \label{edu}
\end{figure}

Addressing academic dishonesty involving AI chatbots like ChatGPT involves recognizing and countering several key challenges. One significant concern is the potential misuse of these chatbots for direct answer retrieval during exams. Students can ask the chatbot questions exactly as they appear in their exams, with the hope of receiving correct answers from the chatbot. By doing this, they can avoid the process of genuinely acquiring knowledge or understanding the subject matter, as they are simply seeking the answers to pass their exams without a deeper understanding of the material. Furthermore, there is a risk associated with essay generation. These chatbots can quickly produce coherent essays on various topics, potentially allowing students to submit well-structured essays even when they lack in-depth understanding or writing skills in the subject matter. Translation assistance is another vulnerability, particularly in language-based assessments. Students can input exam content in one language and rely on the chatbot for translations, thereby accessing information they may not understand in the original language. One notable aspect contributing to these vulnerabilities is the concept-based nature of AI chatbots. When students provide a concept or topic, chatbots generate responses based on similar concepts found in their training data, opening the door to relevant information that may not have been explicitly studied.

AI monitoring for academic integrity goes beyond tracking students' online activities during exams. It involves a comprehensive analysis of behavior to identify suspicious actions. This includes monitoring screen activity for things like switching between applications, tracking mouse and keyboard inputs for irregular patterns, and even using eye-tracking to check if students are reading external content. The system also looks at clipboard activities to spot attempts at copying from external sources. It analyzes response times, flags quick or excessively slow answers, watches for the use of multiple browsers or applications, and can recognize patterns resembling AI-generated content, such as chatbot responses. This approach ensures the detection of potential academic integrity violations, maintaining the exam's integrity. Instructors can enhance assessments by introducing open-ended problems, randomized questions, and adaptive testing, fostering a more personalized and equitable assessment environment and promoting deeper learning. In Figure \ref{edu} we propose a conceptual framework that may help in assessing fair exams with the help of AI based models where the students activity can be monitored for any inappropriateness.

\subsection{Publishing portals}

\begin{figure}
  \centering
    \includegraphics[width=0.75\textwidth]{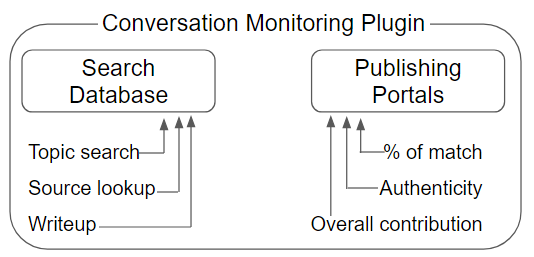}
    \caption{A conceptual architecture for publication portal to asses a submission based on various factors.} 
    \label{pub}
\end{figure}

In the realm of open-access journals, it is crucial to emphasize the significance and essential nature of publications. These publications serve as vital sources of knowledge for a wide range of users, including researchers, groups, and domain experts. Researchers and numerous other publishers depend on peer-reviewed publications as valuable references to enhance their own research endeavors. If publications were to be created with fabricated data and misleading information, it would cast doubts on the quality of research work, posing substantial risks to the research industry.

ChatGPT-like models should come equipped with unique identities or tags for each conversation or generated answer. This feature would enable the content to be easily identified, thereby preventing or halting any fraudulent activities when necessary. For instance, as illustrated in Figure \ref{pub}, we present an architecture where monitoring plugins can be seamlessly integrated with publishing portals to verify the authenticity of content generated using a language model. Additionally, the matching percentage could be thoroughly assessed, considering various factors, and a decision could be made to accept or reject it for initial review. While publishing portals can establish their own safeguards, the provision of such features should be embedded within the content generation platform, such as ChatGPT.

\subsection{Misuse monitoring}

\begin{figure}
  \centering
    \includegraphics[width=0.75\textwidth]{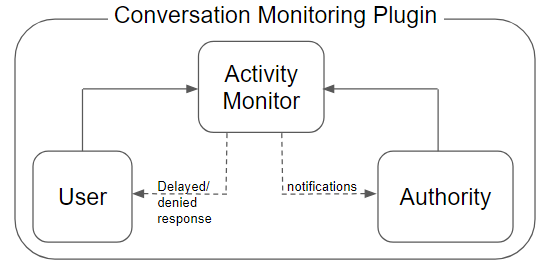}
    \caption{A conceptual architecture to monitor and take action against inappropriate behavior by specific users.} 
    \label{inap}
\end{figure}

ChatGPT has the potential to offer valuable resources, but it's crucial to ensure that these resources are used appropriately, especially when certain age groups, like teenagers, might misuse them on dating sites or for cheating in exams using Language Models (LLMs). While the model's capabilities are vast, maintaining appropriateness is equally significant.

A beneficial approach would involve ChatGPT providing resources only to closed groups, such as registered adults with verified identities or students engaging within an accepted level of conversation with AI-powered chatbots, as defined by authorities. Authoritative access might be necessary in some cases. For instance, teenagers could have access to chatbots connected to their parents, allowing parental notification and the ability to restrict access when inappropriate content arises. This way, specific age groups can learn what's right and wrong. Figure \ref{inap} illustrates an example of how an activity monitoring plugin can be integrated with authorized users to oversee and control activities when needed.

\subsection{Reflections}
ChatGPT should prioritize context when responding to queries, as understanding user perspectives leads to better outcomes. For instance, in subsection \ref{hist}, ChatGPT's agreement about a user's mention of ``another wife" was confusing. It is unclear whether ChatGPT is learning from users, pretending to be right, or assessing right and wrong answers for model improvement. The pattern of responses can sometimes lead to unwanted situations. To address this, ChatGPT should revise its response policies and restrict inappropriate queries.

In the realm of education, institutions should adopt comprehensive strategies and implement AI models with their own policies. This helps ensure that online interactions with ChatGPT align with educational goals and genuinely reflect students' abilities. We recommend that publishing platforms follow a similar approach to maintain research integrity.

ChatGPT has gained widespread acceptance due to its user-friendly nature, and its usage won't decline. Therefore, it's important for models like ChatGPT to prioritize fairness and usefulness for all user groups. We also suggest the inclusion of collaborative chat features for teamwork and the ability to export chats in various formats like PDF, Word, text, or LaTeX. Recognizing and following links, providing a search function within conversations, and offering auto-fill recommendations as users type questions can enhance productivity and improve user understanding of ChatGPT's thought process, enabling them to rephrase questions effectively.

\section{ChatGPT Benefits and Drawbacks}
\label{usage}
In today's digital world, ChatGPT stands as a game-changer, reshaping how we find and use information. Unlike traditional search engines that hand out web links, ChatGPT engages in conversations, offering quick and precise answers. It is more like chatting with a human. What makes it truly unique is its ability to remember what we have talked about before, making interactions feel natural. People are increasingly turning to such AI powered bots because it does not just give facts, it understands context and provides comprehension.

We discuss the benefits and risks of ChatGPT within a limited scope based on our experiment. It may come with extensive opportunities and major risks that are not experimented or discussed in this work.

\subsection{Benefits of ChatGPT}

\textbf{Quick and Precise Reply:} ChatGPT provides very quick answers for queries in a precise manner. It can provide fabricated texts such as bullet points, numbers, and bold formatting, among others.

\textbf{Learning Aid:} ChatGPT can assist in education. It can be a learning companion for any learners with a wide variety of choices. It can simplify complexity and explain in plain, simple language. It can also describe core concepts in a more understandable manner and help students with their studies. The interactive and conversational nature of ChatGPT creates an engaging learning experience, enhancing comprehension and knowledge retention.

\textbf{Coding Assistance:} ChatGPT is great in coding and debugging. It is also an expert in explaining the code, functions line by line, which helps a learner to understand and break down complex code into easier-to-understand steps. Developing new code is easy with ChatGPT.

\textbf{Quick Deployment:} ChatGPT helps in quick development as well. It can generate code swiftly for building a website using basic HTML, CSS, or Markdown, and can also assist with complex structures involving JS, ASP, PHP, etc., as an example.

\textbf{Collaborative Work:} ChatGPT contributes significantly to collaborative knowledge building. For example, it can assist in brainstorming with multiple users facing the same problem. The point is, asking the same question in many different ways brings out various possibilities, making it easy to find the best answer and discard incorrect replies.

\textbf{Multilinguality:} One of ChatGPT's standout features is its multilingual capability. It is able to translate from various languages and transform text into paragraphs as desired. This saves time and enhances productivity, especially for swift documentation.

\textbf{Grammar and Readability:} ChatGPT has the capability to correct grammar and ensure readability if asked, by providing multiple paragraphs. This feature is useful when preparing long documents.

\subsection{Risks of ChatGPT in general}
ChatGPT can come with potential risks while it offers numerous benefits. The main concerns in this report are trust and ethical issues. In this work, we dig deeper to understand how it behaves in general and how it behaves when we try to have a conversation by creating a situation. Having a conversation with ChatGPT may provide us with more results compared to general queries.

\textbf{Ready-Made Answers:} In our experiment, ChatGPT provides ready-made answers for generic questions. We tried with different user accounts from different locations but received pretty much the same answers, leading us to believe that it has ready-made responses.

\textbf{Misinformation:} ChatGPT tries to provide accurate information; however, it is not immune to false or inaccurate information. For example, historical events that are not within the USA region are prone to more errors. This might be due to the less information it has processed in those areas. While it has excellent technical abilities, it suffers from inaccuracies in other types of information, such as places, wars, country-specific details, etc.

\textbf{Potential for Garbage Content:} The ease of content generation with ChatGPT can inadvertently contribute to the production of low-quality or misleading content. Users may misuse ChatGPT to generate articles, reports, or information that lack accuracy or authenticity, which can be detrimental to the credibility of online content.

\textbf{Misuse in Academia and Research:} In academic and research settings, there is a risk that ChatGPT-generated content may be used inappropriately. This includes the submission of ChatGPT-generated work as one's own, which undermines academic integrity and research ethics.

\textbf{Deployment of Fake Services:} Malicious actors may misuse ChatGPT to create fraudulent customer service bots, chatbots, or other automated systems with the intent to deceive or scam users. This can harm both individuals and organizations by eroding trust and causing financial losses.

\textbf{Bias and Fairness:} Another important consideration is the potential bias in ChatGPT's responses. ChatGPT's training data may contain biases present in the broader dataset, leading to biased or unfair responses. Efforts to mitigate bias are ongoing, but users should remain vigilant and critically assess the information they receive.

\textbf{Ethical Dilemmas:} Ethical issues may arise in how ChatGPT is used, particularly in cases where it generates content that promotes harm, hate speech, or misinformation. Decisions about the responsible use of ChatGPT and the content it generates should be guided by ethical considerations to ensure a positive impact on society.

In summary, while the capabilities are remarkable, each user must remain vigilant to its potential drawbacks, ethical concerns, and biases. Anyone may produce any fake things that are not yet identifiable and are prone to greater risks. It is not only challenging from a single-user perspective, but it can also be dangerous for a group of people such as celebrities, athletes, startups, and so on.

\section{Conclusion}
\label{conc}
Throughout this study, we examined the powerful AI-based conversational model across various use cases. It is evident that this tool can be extremely useful in many instances; however, we are also convinced that it can be exploited in certain use cases and become vulnerable when targeted at specific users. Our primary goal was to highlight these limitations in the context of ethical concerns, with the ultimate aim of enhancing ChatGPT's long-term effectiveness. 

This is a part of ongoing research where we further aim to analyse the responses by ChatGPT and other relevant AI chatbots who has the similar approach. At the same time we want to find out how LLMs can learn from the users input, updates its knowledge base accordingly and gain the ability to correct itself when necessary.

\end{document}